\documentclass[conference]{IEEEtran}
\IEEEoverridecommandlockouts
\usepackage{cite}
\usepackage{amsmath,amssymb,amsfonts}
\usepackage{algorithmic}
\usepackage{graphicx}
\usepackage{textcomp}
\usepackage{xcolor}
\usepackage{bbm}
\usepackage{caption}
\usepackage{subcaption}
\usepackage[normalem]{ulem}
\useunder{\uline}{\ul}{}
\def\BibTeX{{\rm B\kern-.05em{\sc i\kern-.025em b}\kern-.08em
    T\kern-.1667em\lower.7ex\hbox{E}\kern-.125emX}}
\begin{document}

\title{Real-World Multi-Domain Data Applications for Generalizations to Clinical Settings \\
}

\author{\IEEEauthorblockN{Nooshin Mojab}
\IEEEauthorblockA{\textit{Department of Computer Science} \\
\textit{University of Illinois at Chicago}\\
Chicago, IL, USA \\
nmojab2@uic.edu}

\\

\IEEEauthorblockN{Manoj Prabhakar Nallabothula}
\IEEEauthorblockA{\textit{Department of Ophthalmology} \\
\textit{University of Illinois at Chicago}\\
Chicago, IL, USA \\
mnalla2@uic.edu}

\\

\IEEEauthorblockN{Joelle A. Hallak}
\IEEEauthorblockA{\textit{Department of Ophthalmology} \\
\textit{University of Illinois at Chicago}\\
Chicago, IL, USA \\
joelle@uic.edu}

\and

\IEEEauthorblockN{Vahid Noroozi}
\IEEEauthorblockA{\textit{Department of Computer Science} \\
\textit{University of Illinois at Chicago}\\
Chicago, IL, USA \\
vnoroo2@uic.edu}

\\

\IEEEauthorblockN{Abdullah Aleem}
\IEEEauthorblockA{\textit{Department of Ophthalmology} \\
\textit{University of Illinois at Chicago}\\
Chicago, IL, USA  \\
aaleem2@uic.edu}

\and

\IEEEauthorblockN{Darvin Yi}
\IEEEauthorblockA{\textit{Department of Ophthalmology} \\
\textit{University of Illinois at Chicago}\\
Chicago, IL, USA \\
dyi9@uic.edu}

\\

\IEEEauthorblockN{Philip S. Yu}
\IEEEauthorblockA{\textit{Department of Computer Science} \\
\textit{University of Illinois at Chicago}\\
Chicago, IL, USA \\
psyu@uic.edu}

}

\maketitle

\begin{abstract}
With promising results of machine learning based models in computer vision, applications on medical imaging data have been increasing exponentially. However, generalizations to complex real-world clinical data is a persistent problem. Deep learning models perform well when trained on standardized datasets from artificial settings, such as clinical trials. However, real-world data is different and translations are yielding varying results. The complexity of real-world applications in healthcare could emanate from a mixture of different data distributions across multiple device domains alongside the inevitable noise sourced from varying image resolutions, human errors, and the lack of manual gradings. In addition, healthcare applications not only suffer from the scarcity of labeled data, but also face limited access to unlabeled data due to HIPAA regulations, patient privacy, ambiguity in data ownership, and challenges in collecting data from different sources. These limitations pose additional challenges to applying deep learning algorithms in healthcare and clinical translations. In this paper, we utilize self-supervised representation learning methods, formulated effectively in transfer learning settings, to address limited data availability. Our experiments verify the importance of diverse real-world data for generalization to clinical settings. We show that by employing a self-supervised approach with transfer learning on a multi-domain real-world dataset, we can achieve $16\%$ relative improvement on a standardized dataset over supervised baselines.   

\end{abstract}


\section{Introduction}

With successful applications of deep neural networks in computer vision, there has been a growing interest in utilizing these machine learning based models on real-world data, particularly in medical imaging. However, the applications on real-world data are limited and challenging. Most of the existing deep learning based approaches have shown promising results by relying on the availability of standardized and adequately labeled datasets. However, real-world data is characterized by variability in quality, machine-type, setting, and source, lacking standardization and labels. Therefore, with the existence of complex real-world data across different clinical applications, and the significant growing interest in employing deep learning models in medical imaging, it becomes crucial to assess the generalization capacity of such models on real-world data applications.  

In this paper, we aim to assess the capacity of deep learning models in coping with real-world data and their generalization aspect by answering two main questions: (1) How well can deep learning based models cope with the complexity of real-world data comprised of multiple device domains versus standardized datasets with a single device domain? And (2) What is the role of real-world data in generalizing to a clinical setting? We answer these two questions by applying deep learning based models on real-world ophthalmic imaging data for the task of glaucoma detection.

Glaucoma is a complex disease that gradually leads to optic nerve damage, resulting in progressive irreversible vision loss. Over 60 million people are diagnosed with glaucoma, encompassing more than 8 million cases with irreversible blindness \cite{davis2016glaucoma}. The global incidence of glaucoma is anticipated to increase up to 111.8 million by 2040 \cite{tham2014global}. 

\begin{figure}[ht]
    \centering
    \includegraphics[width=0.40\textwidth]{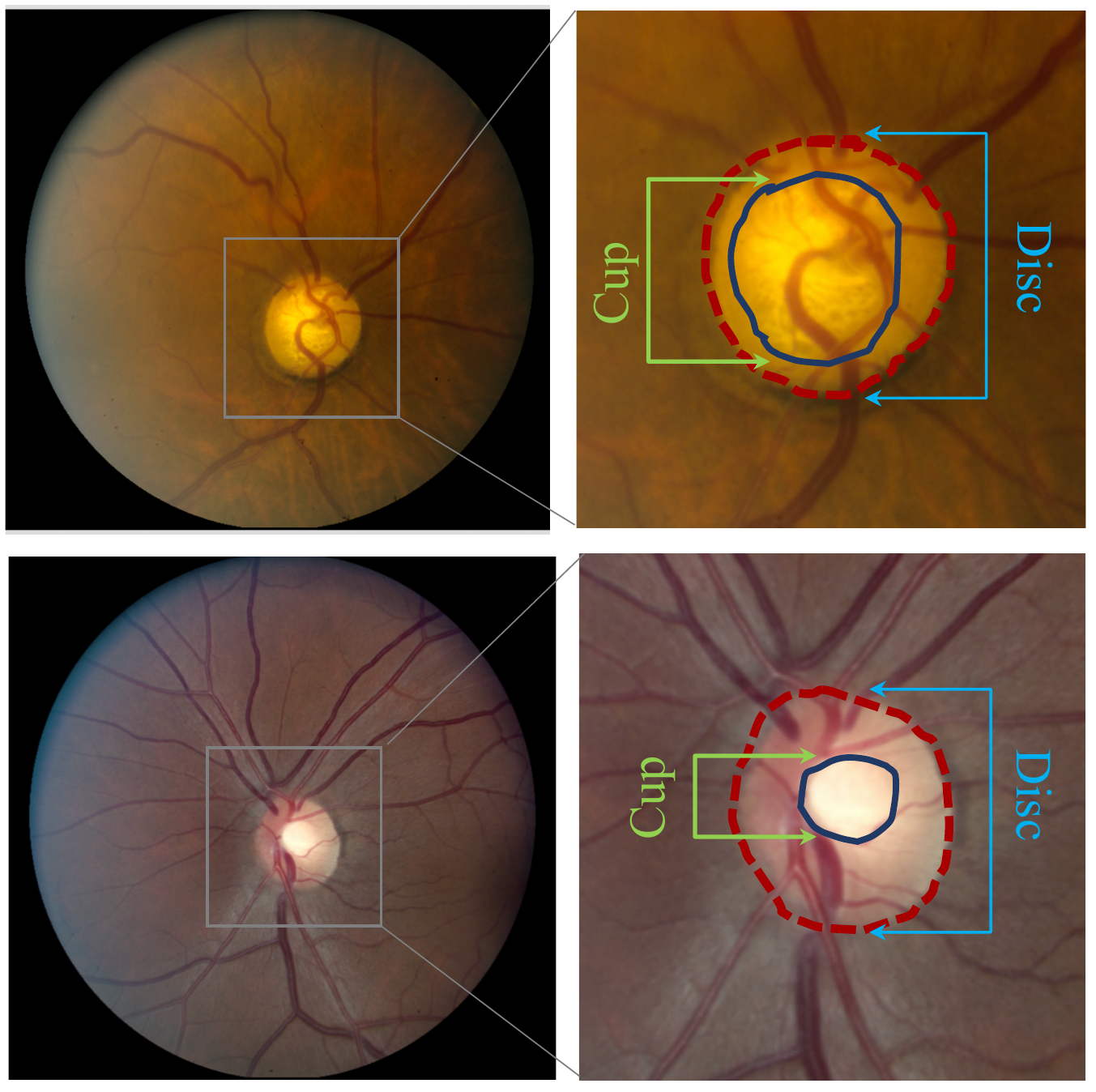}
    \caption{Samples of full Fundus images from I-ODA dataset. The top row represents a glaucomatous Fundus of the eye and the bottom row shows a non-glaucomatous Fundus. The area of the optic nerve head is zoomed in for better visualization of optic disc and cup. CDR is calculated by dividing the cup height (in green arrows) over disc height (in bright blue arrows).}
    \label{fig:cdr}
\end{figure}

Ocular imaging is one of the main modalities used for glaucoma screening. Among the imaging modalities most commonly used for glaucoma, digital Fundus photos are heavily utilized, given their ease and visualization abilities of the disc and cup regions for noninvasive evaluation of the optic nerve head. Assessment of the optic nerve head is based on measuring the optic disc and optic cup regions in Fundus photos and calculating the cup-to-disc ratio (CDR). A sample of glaucomatous and non-glaucomatous optic nerve, with localization of the optic disc and optic cup regions is illustrated in Fig.~\ref{fig:cdr}. All types of glaucoma involve glaucomatous optic neuropathy, and structural progression is characterized by optic nerve thinning, resulting in larger CDR measurements. Glaucoma is suspected when CDR is above some threshold, usually $0.6$. 

Most approaches used to detect the presence of glaucoma in Fundus photos fall into two main categories. (i) Segmentation based approaches utilizing the CDR measurement to detect glaucoma. These approaches involve localizing the disc and cup regions and identifying the presence of glaucoma by measuring the CDR in Fundus photos \cite{khan2013detection},\cite{sevastopolsky2017optic},\cite{zilly2015boosting}. These methods require labeled segmented data requiring the expertise of at least two ophthalmologists to manually mark the region of interest (ROI) in the image. The high variability among graders and low reproducibility limit the application of such approaches. There are some approaches that propose to address the scarcity of segmented labeled data utilizing a multi-task framework \cite{mojab2019deep}. However, their optimal performance is achieved when labeled data for both tasks are available from the same domain. (ii) Classification based approaches in which Fundus photos are being directly fed to a neural network for detecting glaucoma \cite{chen2015glaucoma}, \cite{abbas2017glaucoma}, \cite{raghavendra2018deep}. These approaches mostly employ specifically designed Convolutional Neural Networks (CNNs) trained on small-scaled standardized datasets, limiting their applications to other clinical settings. 

Acquiring labeled data in healthcare is an ongoing challenge as the labeling process not only is time-consuming and expensive but also requires at least two expert graders. Additionally, medical applications suffer from limited access to even unlabeled data due to HIPAA regulations, patient privacy, and ethical considerations, data ownership, and challenges in collecting data from different sources. 

Transfer learning and self-supervised methods have shown promising results in exploiting unlabeled data to learn visual representations \cite{bengio2012deep}, \cite{weiss2016survey} in applications with data limitations such as healthcare. Although self-supervised learning could be a potential solution for applications with data shortages, most of the proposed methods rely on specific design choices for network architecture and predictive pretext tasks \cite{doersch2015unsupervised}, \cite{noroozi2016unsupervised}, \cite{gidaris2018unsupervised}, \cite{zhang2016colorful}. Clinical applications such as glaucoma detection, mostly involve complex multi-domain datasets which limit the employment of such methods that are specifically designed for a particular task or dataset. The recent approach proposed in \cite{chen2020simple}, which is currently the state-of-the-art in self-supervised learning attempts to avoid the complexity of specific design choices by incorporating a broad family of augmentation and contrastive loss into a simple off the shelf architecture, ResNet-50. The composition of data augmentation incorporated in a simple design choice in \cite{chen2020simple} could lead to learning more generalizable visual features and potentially generalize better to other settings. This inspired us to apply the proposed method in \cite{chen2020simple} on a real-world clinical application, glaucoma detection. We believe that real-world clinical applications with diverse multi-domain datasets can benefit from a broad composition of augmentations embedded in a simple framework. This could potentially lessen the sensitivity of the model to domain-specific information in data to some degree and hence improves upon the generalization capacity of the model.

We propose to formulate our problem into a transfer learning framework where we employ the self-supervised approach proposed in \cite{chen2020simple} for visual feature extraction to both alleviate the scarcity of the data and improve generalizations. We evaluate our work by performing extensive experiments on the task of glaucoma detection using real-world datasets. Due to the lack of publicly available large and diverse datasets, we employ a subset of a private clinical ophthalmology dataset referred to as I-ODA which is a diverse multi-domain dataset capturing the complexity characteristic of a real-world data. 

The main contribution of our work is as follows:
\begin{itemize}
    \item We effectively apply the self-supervised representation method via transfer learning on multi-domain real-world data applications and show its superiority over supervised approaches.
    \item We show that leveraging self-supervised visual representation learning via transfer learning could be a potential solution to the limited data in healthcare applications. 
    \item Our experiments shows that deep neural networks have a limited capacity in coping with real-world datasets compared to a standardized dataset.
    \item Our experiments demonstrate that training deep learning models with diverse real-world data generalizes better to clinical settings. %
\end{itemize}

\section{Method}\label{sec:methods}

\subsection{Problem Formulation}\label{ssec:problem_def}
The training data in our problem is comprised of $M$ samples denoted as ${\mathcal{D}} = \big\{(x_k,y_k)| y_k \in \{0,1\} \big\}_{k=1}^{M}$. $y_k$ indicates the binary label of input $x_k$ where values of $1$ and $0$ represent diseased and non-diseased class respectively. Given $\mathcal{D}$, our goal is to learn a binary classifier function $f_c:\cal{X}\xrightarrow{}\cal{Y}$ parameterized by $\theta_c$. We define the following functions 
\begin{equation}\label{eq:base_encoder}
f(x_k;\theta_f)=z_k
\end{equation}
\begin{equation}\label{eq:decoder}
g(z_k;\theta_g)=y_k
\end{equation}

where $f(.)$ parameterized by $\theta_f$ represents the encoder function $f:\cal{X}\xrightarrow{}\cal{H}$ mapping the given input to the latent feature encoding $z_k$. $g(.)$ parameterized by $\theta_g$ represent the decoder function $g:\cal{H}\xrightarrow{}\cal{Y}$ mapping the feature encoding $z_k$ to the label space. Given the input $x_k \in \cal{X}$, function $f_c$ can be decomposed such that 
\begin{equation}\label{eq:decoder_clf}
f_c(x_k, \theta_c)=(g\circ f)(x_k)
\end{equation}
where $\theta_c=\{\theta_g, \theta_f\}$. Given the input $x_k$, binary classifier $f_c(.)$ estimates the probability of an input image being diseased. Both encoder and decoder functions $f(.)$ and $g(.)$ are modeled using neural networks. In particular, the encoder function $f(.)$ is modeled by a Convolutional Neural Network proposed in \cite{chen2020simple} and function $g(.)$ is taken to be a simple linear layer. The loss function of the model comprises a contrastive and classification loss which will be explained in the following sections.

\subsection{Contrastive Loss}\label{ssec:contrast_loss}
To model our encoder function $f(.)$ we employ the approach proposed in \cite{chen2020simple} in which the representations are learned via contrastive loss in the latent space of two augmented views of the same data input. More specifically, this approach comprises three key components: (i) Given the input $x_k$, two augmented views of $x^{(1)}_k$ and $x^{(2)}_k$ are obtained by applying stochastic augmentations, (ii) The base encoder $f(.)$ maps the augmented inputs $x^{(1)}_k$ and $x^{(2)}_k$ to feature maps $z^{(1)}_k$ and $z^{(2)}_k$, (iii) A projection head $q(.)$ maps the feature maps $z^{(1)}_k$ and $z^{(2)}_k$ to latent vectors $h^{(1)}_k$ and $h^{(2)}_k$ respectively. 
Encoder function $f(.)$ and projection head $q(.)$ are both modeled by neural networks. Given two latent vectors $h^{(1)}_k$ and $h^{(2)}_k$, the contrastive loss is defined as 
\begin{equation}\label{eq:contrastive_loss}
\ell_{(1),(2)} = -log \frac{exp(sim_{(1),(2)})/\tau}{\sum_{i=1}^{2N} \mathbbm{1}_{[(i)\neq (1)]} exp(sim_{(1),(i)})/ \tau}
\end{equation}
where $sim_{(i),(j)}$ represents the cosine similarity and $\tau$ represents a temperature parameter. Given a batch of $N$ samples, the application of the two augmentations results in $2N$ samples where pairs of $(h^{(1)}_k,h^{(2)}_k)$ and $(h^{(2)}_k,h^{(1)}_k)$ are accounted as positive samples, and the remainder of pair samples in the augmented batch are accounted as negative. Therefore the overall loss is defined as
\begin{equation}\label{eq:total_contrastive_loss}
\mathcal{L} = \frac{1}{2N} \sum_{i=1}^{N} [\ell((2i-1),(2i)) + \ell((2i), (2i-1))]
\end{equation}

\subsection{Classification Loss}
Given the training set of $\mathcal{D}$, the loss function over the $M$ samples is as follows
\begin{equation}
\label{eq:loss_pred_allData}
\mathcal{L}_c({\mathcal{D}}; \theta_c) =\sum_{k=1}^{M} \ell_c(x_k; \theta_c)
\end{equation}
where $\ell_c(x_k)$ represents the classification loss for sample $x_k$ and it is defined as cross-entropy between the model's estimation and the ground-truth label
\begin{equation}
\label{eq:cross_entropy_loss}
\ell_c(x_k;\Theta_c) = -y_k \log {\hat{y}_k} - (1 - y_k) \log (1-\hat{y}_k))
\end{equation}
where $\hat{y}_k=f_c(x_k;\Theta_c), 0<\hat{y}_k<1$ indicates the model's estimation for sample $x_k$ being diseased or not and $y_i$ represent its corresponding ground-truth label. 
We aim to perform the task of detecting glaucoma disease given the Fundus photo. Therefore our dataset is a collection of Fundus photos labeled as either glaucoma ($y_i=1$) or non-glaucoma ($y_i$=0).

\subsection{Transfer Learning}
We propose the solution to our task by formulating our problem into the transfer learning framework. For the purpose of this paper, we employ the pretrained encoder networks obtained via the approach proposed in \cite{chen2020simple}, where the projection head is thrown away and the encoder function $f$ and feature maps $z_k$ are used to perform the binary classification task of glaucoma detection. We take the base encoder $f$ to be ResNet-50 as suggested in \cite{chen2020simple} and consider two settings for transfer learning across our datasets: (i) using the fixed features extracted from a pretrained encoder network to train a linear classifier on top of the frozen based network, and (ii) fine-tuning a fraction of the network. In this scenario, we utilize the weights of the pretrained network as initialization. 

\section{Experiments}

\subsection{Dataset}\label{sec:oda_dataset}
\begin{figure}[ht]
    \centering
    \includegraphics[width=0.40\textwidth]{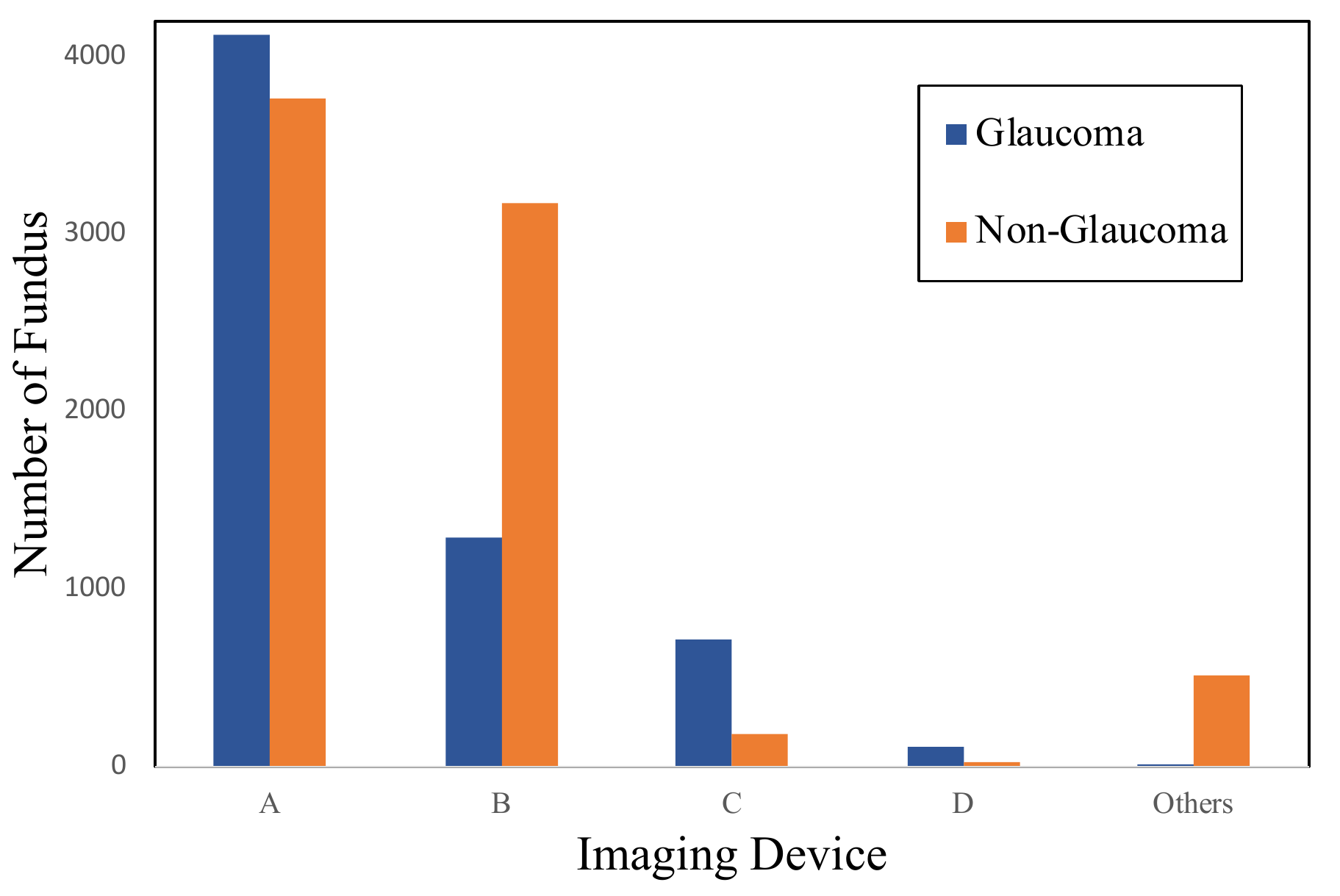}
    \caption{The number of Fundus photos per imaging device. The majority of Fundus images are generated by devices A and B.}
    \label{fig:oda_stat}
\end{figure}

\noindent\textbf{I-ODA} \footnote{For questions related to accessing the I-ODA database please contact the author, Joelle Hallak.} has been created from imaging data belonging to patients who visited the Illinois Eye and Ear Infirmary of the University of Illinois Chicago (UIC). This dataset was created by assigning $\sim 4M$ unlabeled images into $12$ categories based on the imaging devices by utilizing metadata information and a hybrid approach of machine learning algorithms and manual search. After labeling images into one of the $12$ categories, patients were labeled with their corresponding diagnosis using billing information (ICD codes). We isolated $6244$ glaucoma and $7664$ non-glaucoma patients from this dataset. Non-glaucoma patients in our dataset are selected as those patients being diagnosed with neither glaucoma, nor glaucoma suspect with no potential damage to the optic nerve head. Among the $12$ categories, we isolated Fundus images for the purpose of our experiments in this paper which we refer to as ODA-G dataset. Each Fundus image can be generated by a different imaging device. The statistics of image data and the corresponding imaging device distribution in the ODA-G dataset is illustrated in Fig.~\ref{fig:oda_stat}. 
%

\begin{figure*}
\begin{subfigure}{.5\textwidth}
  \centering
  \includegraphics[width=.85\linewidth]{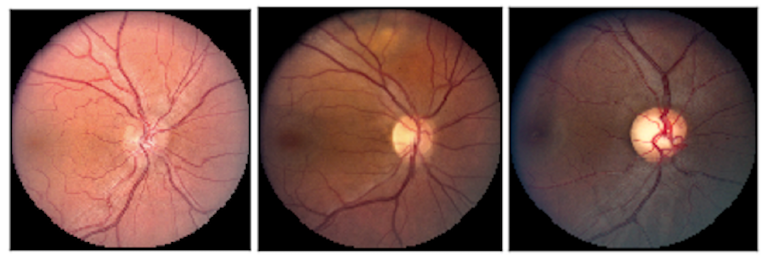}  
  \caption{Fundus samples from device A}
  \label{fig:ois}
\end{subfigure}
\begin{subfigure}{.5\textwidth}
  \centering
  \includegraphics[width=.85\linewidth]{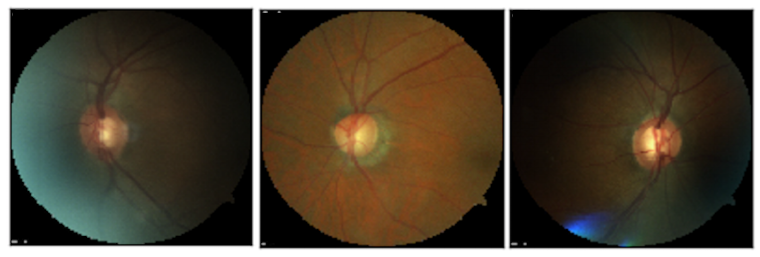}  
  \caption{Fundus samples from device B}
  \label{fig:cirrus}
\end{subfigure}
\newline
\begin{subfigure}{.5\textwidth}
  \centering
  \includegraphics[width=.85\linewidth]{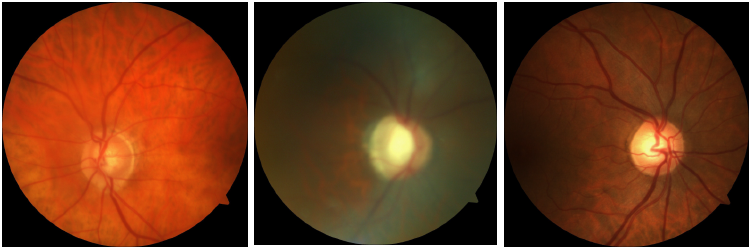}  
  \caption{Fundus samples from device C}
  \label{fig:Fundus}
\end{subfigure}
\begin{subfigure}{.5\textwidth}
  \centering
  \includegraphics[width=.85\linewidth]{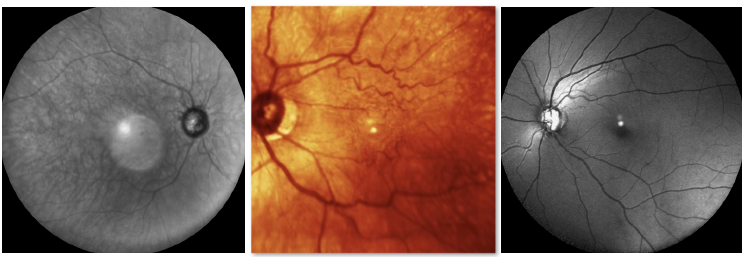}  
  \caption{Fundus samples from device D}
  \label{fig:spectralis}
\end{subfigure}
\caption{A snapshot of samples of Fundus photos generated by different devices. The top row illustrates a subset of images from devices A and B from left to right. The bottom row illustrates a subset of Fundus generated by devices C and D from left to right.}
\label{fig:device_samples}
\end{figure*}

Imaging data in the ODA-G dataset is spread across $11$ different imaging devices where devices A, B, C, and D comprise $96\%$ of the data, while the other $7$ devices are responsible for almost $4\%$ of the data. The data distribution for images generated from each device can be different and hence we can have a dataset which is a mixture of different domains. A snapshot of samples of Fundus photos from $4$ major imaging devices A, B, C, and D are illustrated in Fig.~\ref{fig:device_samples}. 
As can be seen, images across different devices can have different color distributions, different shapes, and positioning of the cup and disc regions. We refer to imaging devices as different domains in our dataset. In order to analyze the complexity of coping with complex real-world data and its generalization capacity, we consider two subsets of data, (i) Fundus images generated from only one imaging device. We isolated the Fundus photos from device A which we refer to as ODA-A and (ii) ODA-G which contains Fundus generated by all devices used for glaucoma screening in our dataset. Both subsets are generated from a real-world setting, however, in this scenario, ODA-A comprises a single domain data resembling the properties of a standardized dataset to some extent. On the other hand, ODA-G comprises multi-domain data capturing the complex aspect of diverse real-world data.   
We aim to answer two main questions using these two subsets, (1) how do deep learning models cope with the complexity of real-world data as opposed to a standardized dataset? (2) Is training the model on real-world data important for generalizations?

Our goal was to also validate our results on an external dataset. However, to the best of our knowledge, the availability of public datasets with adequate samples of full Fundus photos for glaucoma detection is very limited. Therefore we performed all of our experiments on our datasets, ODA-G and ODA-A.

\subsection{Experimental Setting}

\subsubsection{Base Encoder} \label{ssec:base_enc}

The base encoder $f$ and projection function $q$ are modeled with ResNet-50 and $2$ layer MLP with Rectified Linear Unit (ReLU) activation respectively. We employ pretrained networks on ImageNet and CIFAR-10 in our experiments. The CIFAR-10 based network employs a simpler network compared to the ImageNet based network. The CIFAR-10 based encoder employs a ResNet like architecture with a depth of $18$ composing of $4$ main residual blocks each consists of two stacked convolutional followed by batch normalization. The ImageNet based encoder employs ResNet-50 with three hidden layers widths ($1\times$, $2\times$, $4\times$) \cite{chen2020simple}, \cite{kolesnikov2019revisiting} which we utilize in our experiments. When using the ImageNet pretrained network, a random crop and resize with random flip, color distortions, and Gaussian blur are employed for data augmentations. When CIFAR-10 is used, only random crop and resize with random flip and color distortions are employed for augmentation \cite{chen2020simple}.  

\subsubsection{Transfer Learning}
We evaluate our transfer learning setting across two datasets ODA-G and ODA-A by (1) learning a linear classifier on top of the pretrained network and (2) fine-tuning $x\%$ of the network where $x \in \{25, 50, 75, 100\}$. 

In the first setting, a logistic regression classifier is trained on top of the frozen base encoder network. The extracted features from the pretrained network are used to perform binary classification for detecting glaucoma in a given Fundus input. In this scenario, no data augmentation is applied. The number of epochs is selected from $\{50, 100, 200, 300\}$, batch size is selected from $\{1024, 512, 256, 128\}$ and learning rate is selected from $\{0.1, 0.07, 0.05, 0.03\}$.

In the second setting, we fine-tune the base network using the pretrained network's weights as initialization. We fine-tune $\{25\%, 50\%, 75\%, 100\% \}$ of the network for $\{50, 100, 200, 300\}$ epochs where learning rate is selected from $\{0.1, 0.07, 0.05, 0.03, 0.01, 0.007\}$. The batch size is selected from $\{512, 256, 128, 64\}$ and the rest of hyperparameters are set to default value as reported in \cite{chen2020simple}. For optimizer we use SGD optimizer with Nestrov momentum with momentum parameter set to $0.9$ and Adam. 

In all of our experiments, input images were resized to $224\times 224$ when using ImageNet based pretrained network and resized to $32\times 32$ when CIFAR-10 based pretrained network is employed. The split of $80\%$ and $20\%$ is used for training and testing. The hyperparameter tuning for each dataset is performed on the validation set chosen randomly as $20\%$ subset of the training set. After finding the appropriate values for hyperparameters, all the training set is used for training. Notably, each patient in our dataset can have more than one Fundus photo taken in multiple exam sessions. To prevent the leakage of information from the test set into training, we split the data by patients. Eventually, the result on the test data is reported in terms of the accuracy metric.

\subsubsection{Baselines} \label{sec:baseline}
We compare our results with fully supervised methods. Previous approaches proposed for glaucoma detection mostly employ a Convolutional Neural Network (CNN) trained on small unique datasets. Since most of these studies do not release their code publicly, we simulate the employed general approach by developing a CNN model and consider that as our supervised baseline. 
We employ two approaches for our supervised baselines. (1) For the sake of comparison with our proposed approach, we employ ResNet-50, similar to the base encoder network architecture, followed by MLP with $1$ hidden layer of size $1024$, batch normalization, ReLU activation, and dropout. Then sigmoid activation function is applied on the last layer to get the final output. (2) A simpler CNN specifically designed for our dataset. Our CNN network is composed of $2$ convolutional block, where each consists of $1$ convolutional layers with ReLU activation, batch normalization, non-linearity ReLU, and max pooling. Two fully connected layers with size $512$ and $64$ followed by batch normalization, non-linearity ReLU, and dropout are further applied. The Sigmoid activation function is used on the final layer to get the output of the model. 

We consider both random and ImagNet weights as initialization for the ResNet-50 model. We consider our supervised baselines in both absence and presence of data augmentation. We explore color distortion and random flip for our augmentation.

\subsection{Experimental Results}
In this section we attempt to answer our two main questions through extensive experiments performed on the task of glaucoma detection: (1) How well can deep learning based models cope with the complexity of real-world data comprised of multiple device domains versus standardized datasets with a single device domain? and (2) What is the role of real-world data in generalizing to a clinical setting?  

We first demonstrate the effectiveness of combining a self-supervised method with transfer learning on applications with limited data compared to fully supervised models. Further, to address our two main goals, we assess the capacity of neural networks in dealing with complex real-world data by comparing the performance of the model on a multi-domain real-world dataset, ODA-G, as opposed to a single-domain dataset ODA-A, resembling a standardized and simpler dataset. Additionally, we assess the role of learning with real-world data for generalization to a clinical setting (glaucoma detection). 

We also analyze the effect of data augmentation, training time, and fraction of fine-tuning on the performance of the model.

\subsubsection{\textbf{Comparison against fully supervised approach}}
In this section, we demonstrate the result of our work on both ODA-G and ODA-A dataset for the task of glaucoma detection using transfer learning in both linear evaluation and fine-tuning settings.
We first evaluate our approach by learning a linear classifier using self-supervised representations learned by a pretrained network and compare the result with supervised baselines. To potentially improve the performance of the network compared to using the fixed extracted features, we further investigate the performance of our classifier with fine-tuning the base encoder network using the weights of the pretrained network as initialization. We experiment with fine-tuning $\{25\%, 50\%, 75\%, 100\% \}$ of the network. For self-supervised pretrained networks, we consider both ImageNet and CIFAR-10 based networks proposed in \cite{chen2020simple}. For the ImageNet based pretrained network, we report the results for all ResNet-50 ($1\times$, $2\times$, $4\times$) \cite{chen2020simple}, \cite{kolesnikov2019revisiting}. For the CIFAR-10 based network, we report the result on the simpler ResNet. We report the best result achieved for each combination of dataset and model in Table~\ref{tab:linear_ft_eval}. 
\begin{table}
\centering
\caption{Comparison of employing self-supervised learned representations via transfer learning (TR) in linear evaluation and fine-tuning settings against fully supervised baselines.}
\label{tab:linear_ft_eval}
\begin{tabular}{lccc}
\hline
 & ODA-G & ODA-A & Weights \\ \hline
{\ul \textit{Supervised Baselines}} &  & \textbf{} &  \\
ResNet-50 (1$\times$) & 59.93 & 83.57 & Random \\
ResNet-50 (1$\times$) (+Augmentation) & 82.61 & 86.44 & Random \\
CNN & 64.67 & 88.35 & Random \\
CNN (+Augmentation) & 82.93 & 90.43 & Random \\
\hline
{\ul \textit{self-supervised TR+Linear}} &  & \textbf{} &  \\
ResNet  & 72.87 & 82.40 & CIFAR-10 \\
ResNet-50 (1$\times$) & 80.26 & 88.92 & ImageNet \\
ResNet-50 (2$\times$) & 83.84 & 91.00 & ImageNet \\
ResNet-50 (4$\times$) & \textbf{84.30} & \textbf{92.34} & ImageNet \\ \hline
{\ul \textit{self-supervised TR+Fine-tuning}} &  & \textbf{} &  \\
ResNet & 82.05 & 86.04 & CIFAR-10 \\
ResNet-50 (1$\times$)  & 83.28 & 90.35 & ImageNet \\
ResNet-50 (2$\times$) & 83.14 & 90.43 & ImageNet \\
ResNet-50 (4$\times$) & 82.05 & 90.11 & ImageNet \\ \hline
\end{tabular}
\end{table}

The result depicted in Table~\ref{tab:linear_ft_eval} shows the superiority of transfer learning using self-supervised learned representations over fully supervised approaches. We achieve our best result with ResNet-50 (4$\times$). For supervised baselines, we only report the result for ResNet-50 with random initialization as we did not observe any significant change between the two weight initializations (i.e. random and ImageNet weights). We can also see that employing wider networks can further improve the performance of the linear classifier on the test set. 

We observed that fine-tuning the network does not significantly change the performance of the model compared to training a linear classifier. Moreover, the result shown in Table~\ref{tab:linear_ft_eval} indicates that employing ImageNet pretrained weights compared to CIFAR-10 achieves a better result on both datasets ODA-G and ODA-A. This result is expected as training with a more diverse dataset comprised of a broader range of image categories, helps with learning more generalizable features.  

\subsubsection{\textbf{Medical domains benefit from transfer learning using self-supervised learned features}} Among the supervised baselines in Table~\ref{tab:linear_ft_eval}, the CNN (+Augmentation) method that is specifically designed for our dataset-task achieves a better result over off the shelf ResNet networks. This result suggests that specifically designed networks for each dataset-task combination are usually crucial to the success of supervised approaches which limits their applications and generalization capacity to some extent. The superiority of our framework over supervised baselines shows that we can avoid the complexity of design choice for each particular task by simply using one of off the shelf networks. This result is particularly important in medical imaging where applications are limited with data to train robust supervised models. 

\subsubsection{\textbf{Neural Networks On real-world Data}}

In this section, we assess the capacity of neural networks in coping with real-world datasets versus standardized datasets, commonly used by most deep learning algorithms. We analyze our experimental results on the two datasets ODA-G and ODA-A. As a reminder, ODA comprises data from multiple imaging devices forming a complex multi-domain dataset, while ODA-A comprises a single domain data representing a less challenging standardized dataset. Naturally, a mixture of different data distributions emanating from different imaging devices in the ODA-G dataset poses a major challenge for training deep neural networks. From Table~\ref{tab:linear_ft_eval} we observe that experiments on ODA-A significantly outperform the experiments on ODA-G in all three settings supervised, linear evaluation, and fine-tuning. The performance gap is particularly noticeable in supervised settings as their success usually relies on the availability of large standardized labeled datasets. This experiment verifies that deep learning models do not perform as well on complex real-world datasets as on standardized datasets. 
 
\subsubsection{\textbf{Training deep learning models with diverse real-world data generalizes better to clinical settings.}}

To assess the role of training with complex real-world data in generalization to a clinical setting, we design an experiment where we form four pairs of (ODA-G, ODA-G), (ODA-G, ODA-A), (ODA-A, ODA-G) and (ODA-A, ODA-A) datasets. We train on the first element of each pair and evaluate on the second element. The result of this experiment is shown in Table ~\ref{tab:gen_unsup}.
\begin{table}
\centering
\caption{Generalization across pairs of datasets employing self-supervised learned representation via transfer learning.}
\label{tab:gen_unsup}
\begin{tabular}{ccc}
\hline
\multicolumn{1}{l}{Training Data} & \multicolumn{1}{l}{Testing Data} & \multicolumn{1}{l}{Test Accuracy(\%)} \\ \hline
ODA-G & ODA-G & 84.30 \\
ODA-G & ODA-A & {\color[HTML]{009901} \textbf{90.35}} \\
ODA-A & ODA-G & {\color[HTML]{FE0000} \textbf{76.53}} \\
ODA-A & ODA-A & 92.34 \\ \hline
\end{tabular}
\end{table}
As the results suggest if we only train the model on one small unique dataset, ODA-A, and evaluate on an extremely complex real-world dataset, ODA-G, we achieve the worst result depicted in red in Table~\ref{tab:gen_unsup}. On the other hand, when we train the model using a diverse multi-domain real-world dataset, ODA-G and evaluate it on a smaller and more unique dataset, ODA-A, we achieve a promising result as depicted in green in Table~\ref{tab:gen_unsup}. This result indicates that even though complex data makes the learning process harder, it leads the model towards learning more generalizable features. It is not surprising that (ODA-A, ODA-A) experiment achieves the best result, as this is where usually neural networks perform best. However, we can see that the obtained result from the (ODA-G, ODA-A) experiment is also competitive with the result of the (ODA-A, ODA-A) experiment, indicating the advantage of learning with real-world data.
The overall results validate the crucial role of real-world data for generalizing to clinical settings. If the network trains on small unique datasets, it fails to generalize to other domains. Hence we need complex diverse datasets that capture aspects of real-world data to cope with generalizations to other domains, especially in clinical settings.

\textbf{Self-supervised learning approaches generalize better on real-world data.}
We would like to assess the capacity of self-supervised approaches in generalization compared to supervised methods. We perform the same experiments but this time under supervised settings. The result is shown in Table~\ref{tab:gen_supervised}.
\begin{table}
\centering
\caption{Generalization across pairs of datasets employing supervised approaches.}
\label{tab:gen_supervised}
\begin{tabular}{ccc}
\hline
Training Data & Testing data & Test Accuracy (\%) \\ \hline
ODA-G & ODA-G & 82.93 \\
ODA-G & ODA-A &  \textbf{74.00} \\
ODA-A & ODA-G & \textbf{69.19} \\
ODA-A & ODA-A & 90.43 \\ \hline
\end{tabular}
\end{table}
The observation from Table~\ref{tab:gen_supervised} (shown in bold) supports the result in our previous section. Training with a real-world dataset, ODA-G, generalizes better on ODA-A dataset than the reversed experiment (ODA-A, ODA-G). However, we can see that the performance gap between (ODA-G, ODA-G) and (ODA-G, ODA-A) experiment and between (ODA-A, ODA-A) and (ODA-A, ODA-G) is more noticeable compared to the result in Table~\ref{tab:gen_unsup}. This result may indicate that supervised approaches perform the best when they are trained and evaluated on the same dataset. Moreove, the comparison between the results in Tables~\ref{tab:gen_unsup} and ~\ref{tab:gen_supervised} shows that (ODA-G, ODA-A) performs poorly under a supervised approach achieving only $74\%$ accuracy, while under a self-supervised approach it achieves $90.35\%$ which is more than $16\%$ improvement over the supervised setting. This result verifies the superior capacity of self-supervised approaches in generalizing across different device domains, indicating another advantage of applications of self-supervised representation learning methods.

\subsubsection{\textbf{Performance Analysis}}
In this section, we first analyze the role of data augmentation on generalization. Then we evaluate the effect of training time and fine-tuning $x\%$ of the network on the performance. 

\textbf{Data augmentation is crucial for generalization on real-world data.}
To assess the role of data augmentation in generalization, we explore the behavior of our supervised approach in both the presence and absence of data augmentation. For ODA-G we applied the composition of color distortion and random flip and for ODA-A we only applied the random flip. As the result in Table~\ref{tab:linear_ft_eval} suggests, data augmentation can efficiently improve the generalization on the test data. This result shows that incorporating data augmentation enhances the capacity of the model to learn more generalizable features. This improvement is particularly more noticeable for the ODA-G dataset containing multi-domain data. This result could indicate that the model leverages the most from data augmentation when training on a diverse real-world dataset. We believe that the composition of data augmentation utilized in \cite{chen2020simple}, also contributes significantly in the success of this method on our task. 

\begin{figure}[ht]
\begin{subfigure}{.5\textwidth}
  \centering
  \includegraphics[width=.74\linewidth]{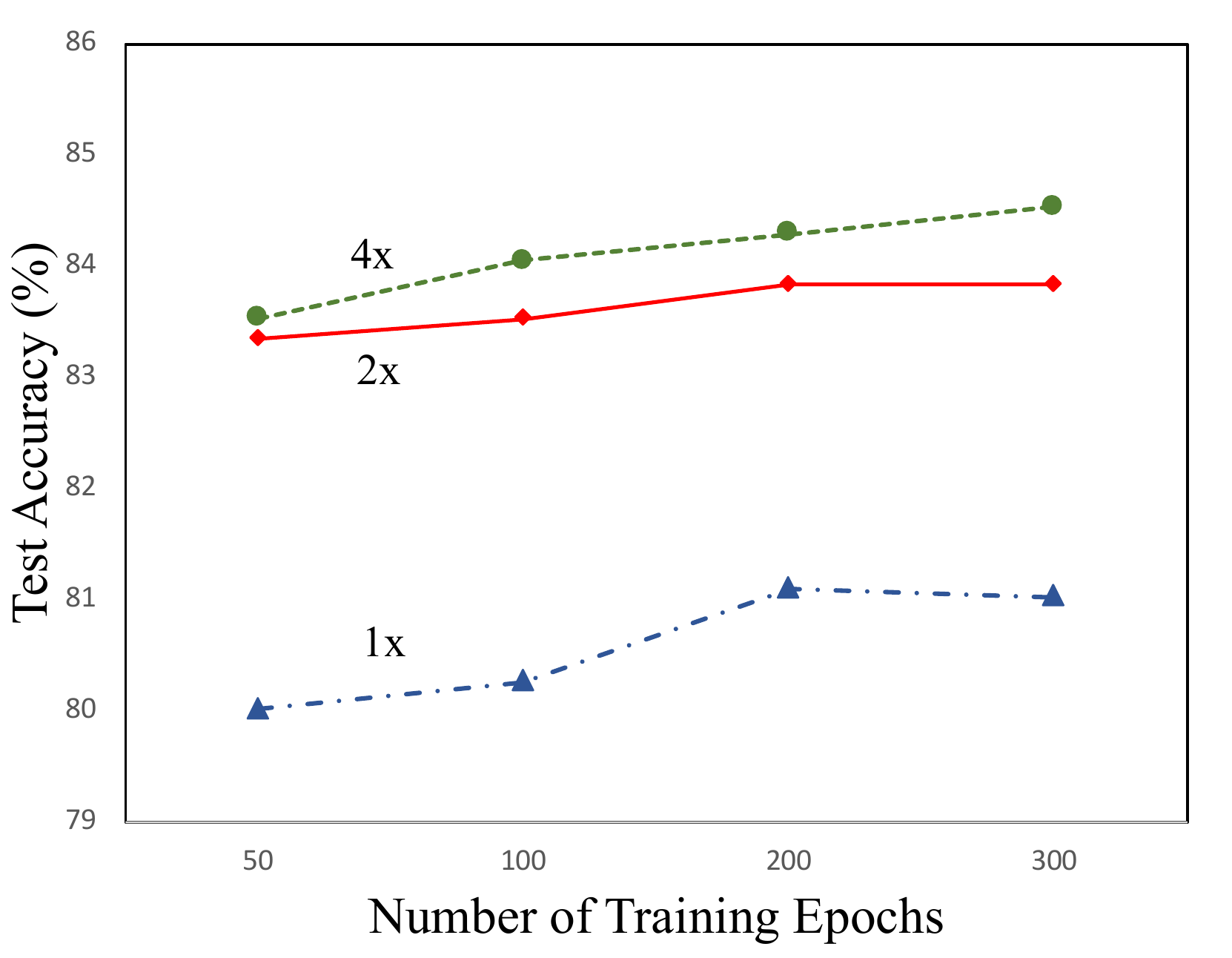}  
  \caption{ODA-G dataset}
  \label{fig:odag_epochs}
\end{subfigure}
\begin{subfigure}{.5\textwidth}
  \centering
  \includegraphics[width=.74\linewidth]{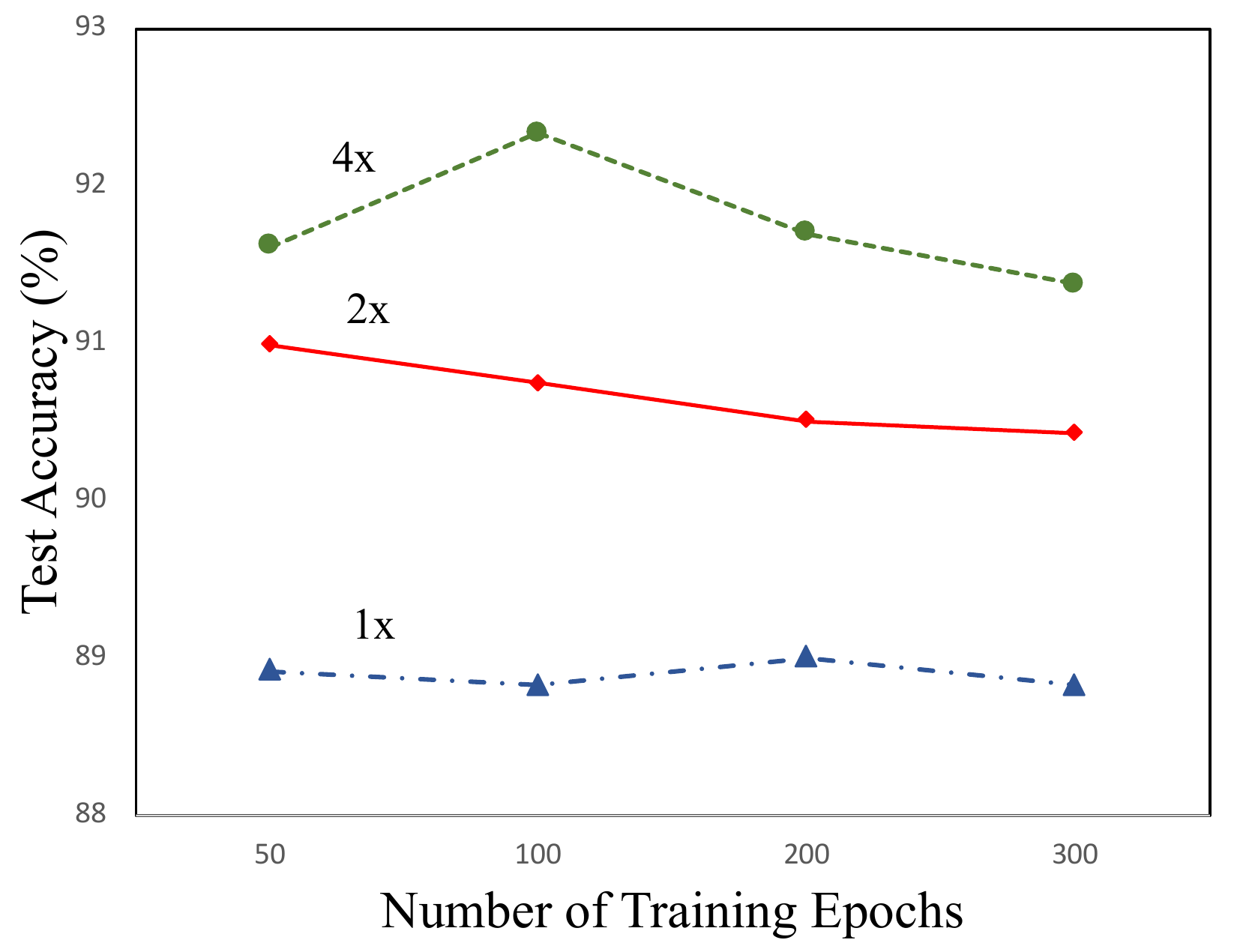}  
  \caption{ODA-A dataset}
  \label{fig:ois_epochs}
\end{subfigure}
\caption{Effect of the number of training epochs on testing accuracy employing ImageNet based pretrained network for ODA-G and ODA-A datasets.}
\label{fig:acc_epoch_imgnet}
\end{figure}

\begin{figure}[ht]
    \centering
    \includegraphics[width=0.36\textwidth]{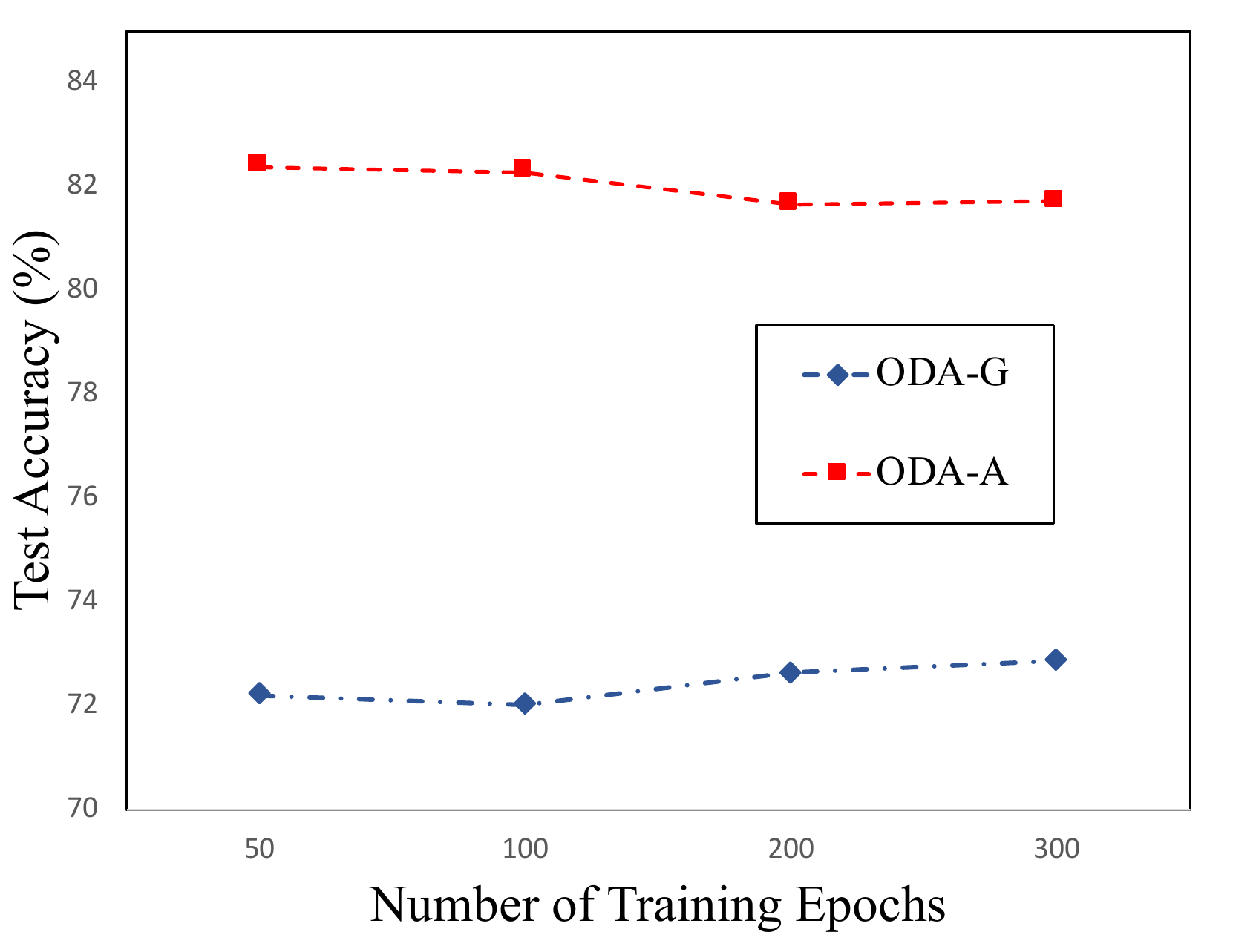}
    \caption{Effect of the number of training epochs on testing accuracy employing CIFAR-10 based pretrained network for ODA-G and ODA-A datasets.}
    \label{fig:acc_epoch_cifar}
\end{figure}

\textbf{Diverse large-scaled datasets benefit from longer training.} We perform our experiments by increasing the number of training epochs while keeping the batch size fixed for each model and report the accuracy on the testing set. Fig.~\ref{fig:acc_epoch_imgnet} and ~\ref{fig:acc_epoch_cifar} shows the plots of test accuracy versus training epochs for linear evaluation employing ImageNet based ResNet-50 ($1\times$, $2 \times$, $4 \times$)  and CIFAR-10 based ResNet respectively on both ODA-G and ODA-A datasets.

%
As the plots in Fig.~\ref{fig:acc_epoch_imgnet} suggest, the performance improves as we increase the number of training epochs when employing the ImageNet based encoder network on the ODA-G dataset. However, increasing the number of epochs has the opposite effect on the ODA-A dataset. We observe a similar effect when using CIFAR-10 based encoder as the Fig.~\ref{fig:acc_epoch_cifar} suggests. However, we can see that increasing the number of epochs has a less effect on performance improvement on ODA-G compared to using the ImageNet based encoder. The overall result indicates that more diverse and larger datasets can benefit more from longer training. 

\begin{figure}[ht]
    \centering
    \includegraphics[width=0.37\textwidth]{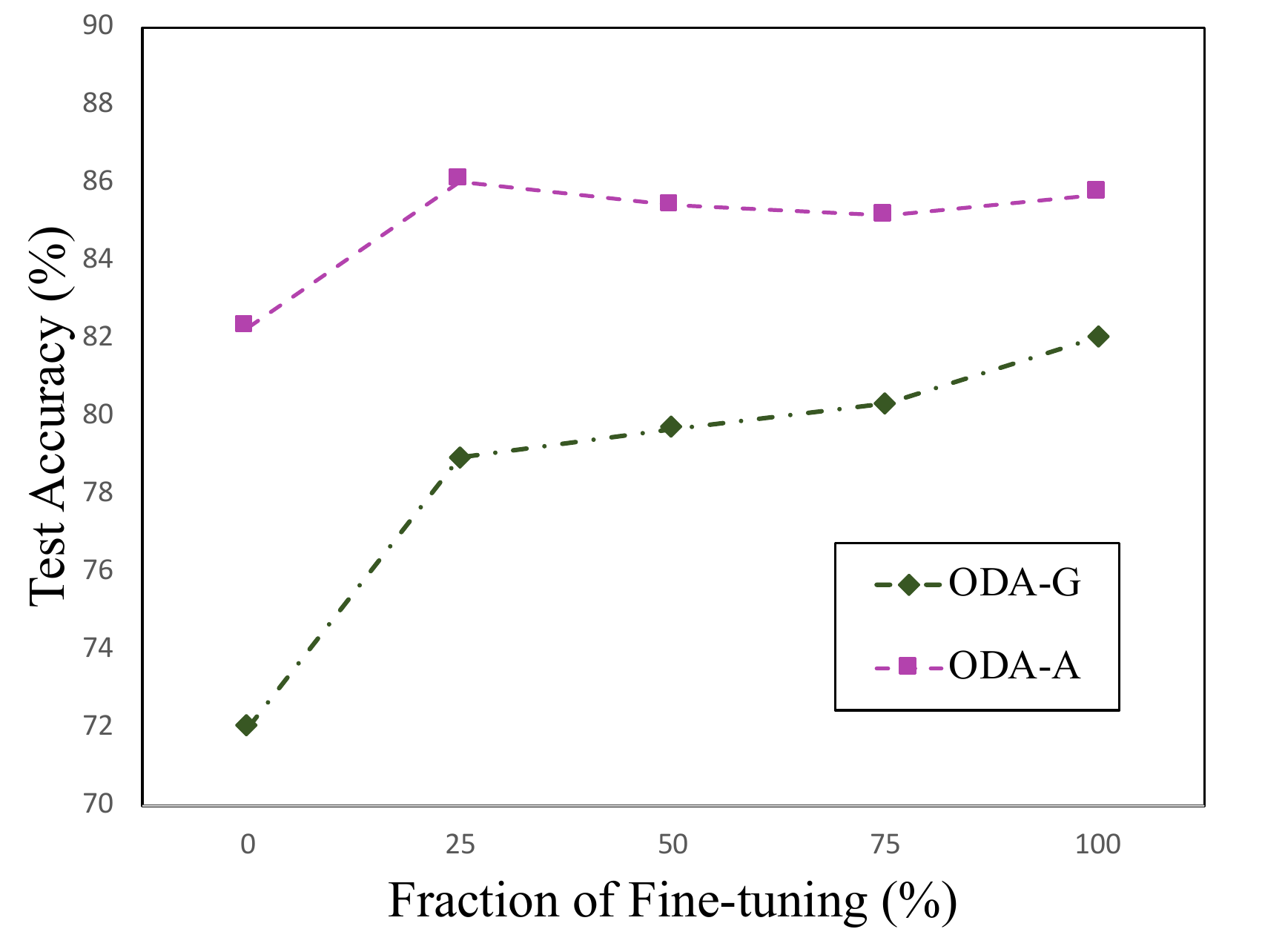}
    \caption{The effect of $x\%, x\in \{25, 50, 75, 100\}$ fine-tuned network on test accuracy using CIFAR-10 based encoder network.}
    \label{fig:ft_percentage}
\end{figure}

Additionally, we plot the effect of fine-tuning $x\%$ of the network on test accuracy. The result is shown in Fig.~\ref{fig:ft_percentage} when fine-tuning the network using the CIFAR-10 based encoder network. 

As the plot in Fig.~\ref{fig:ft_percentage} suggests, the test accuracy improves as we fine-tune a larger portion of the network when training on ODA-G dataset. Training on the ODA-A still benefits from fine-tuning the network especially when it is fine-tuned on $1/4$ of the network but after that, we do not observe a significant change in the performance. We achieved our best result for this experiment by fine-tuning the whole network from scratch when using the ODA-G dataset and $25\%$ of the network when using ODA-A. We also observed that the model does not benefit substantially from fine-tuning the network with ImageNet weights.

\section{Conclusion}
In this paper, we utilized self-supervised visual representation learning methods effectively formulated in transfer learning settings to alleviate the shortage of data in medical imaging applications and improve upon the generalization capacity of the model. We verified our results by performing a glaucoma detection task on a real-world ophthalmic data application. We showcased the importance of learning with real-world data for generalization, through performing extensive experiments on a multi-domain real-world dataset versus a single-domain standardized dataset. Additionally, We showed that without learning on complex real-world data, the deep learning models cannot generalize well to clinical settings.  

\bibliographystyle{IEEEtran}
\bibliography{refs}

\end{document}